\documentclass[conference]{IEEEtran}
\def\IEEEtitletopspace{18pt}

\IEEEoverridecommandlockouts
\usepackage{cite}
\usepackage{amsmath,amssymb,amsfonts}
\usepackage{algorithmic}
\usepackage{graphicx}
\usepackage{textcomp}
\usepackage{xcolor}
\usepackage{mathtools}

\usepackage{array}

\usepackage[mathscr]{eucal}

\newcommand{\etal}{\textit{et al.}}
\usepackage{subfig}
\usepackage{xfrac}

\makeatletter
\newcommand{\linebreakand}{%
  \end{@IEEEauthorhalign}
  \hfill\mbox{}\par
  \mbox{}\hfill\begin{@IEEEauthorhalign}
}
\makeatother

\def\BibTeX{{\rm B\kern-.05em{\sc i\kern-.025em b}\kern-.08em    T\kern-.1667em\lower.7ex\hbox{E}\kern-.125emX}}
\begin{document}

\title{{Integrating occlusion awareness in urban motion prediction for enhanced autonomous vehicle navigation}
\thanks{This work has been partially funded by MCIN/AEI/10.13039/501100011033 and by ERDF A way of making Europe with the project DISCERN (PID2021- 125850OB-I00).}}



\author{\IEEEauthorblockN{Vinicius Trentin\IEEEauthorrefmark{1}, Juan Medina-Lee\IEEEauthorrefmark{2}, Antonio Artuñedo\IEEEauthorrefmark{1} and Jorge Villagra\IEEEauthorrefmark{1}}
\IEEEauthorblockA{\IEEEauthorrefmark{1}Centre for Automation and Robotics (CSIC-UPM), Arganda del Rey, Spain} 
\IEEEauthorblockA{\IEEEauthorrefmark{2}{Computer Science and Engineering Department, University of Puerto Rico, Mayagüez, USA}\\
Email: (vinicius.trentin, antonio.artunedo, jorge.villagra)@csic.es, juan.medina26@upr.edu}}



\maketitle

\begin{abstract}
Motion prediction is a key factor towards the full deployment of autonomous vehicles. It is fundamental in order to ensure safety while navigating through highly interactive and complex scenarios. Lack of visibility due to an obstructed view or sensor range poses a great safety issue for autonomous vehicles.
The inclusion of occlusion in interaction-aware approaches is not very well explored in the literature.
In this work, the MultIAMP framework, which produces multimodal probabilistic outputs from the integration of a Dynamic Bayesian Network and Markov chains, is extended to tackle occlusions. The framework is evaluated with a state-of-the-art motion planner in two realistic use cases.

\end{abstract}


\section{Introduction}

The perception system for autonomous vehicles has made significant advancements in recent years. However, full knowledge of the environment without V2X communication remains unrealistic in real-world situations. The perception of the vehicle will inevitably be limited by the presence of other vehicles, static objects, or sensor range.

Ignoring the possibility of occlusions\footnote{In this paper, the term occlusion is used to refer to areas from where a vehicle might appear.} can be extremely hazardous, as sudden appearances of objects from an occluded zone can occur at any time. This can result in accidents and reckless behavior, especially when navigating through scenarios where there is a high density of objects and vehicles. 


Therefore, it is crucial to design occlusion-aware motion prediction strategies to account for these uncertainties and ensure safe and reliable navigation. As a result, the autonomous driving system (ADS) can better estimate the probability of different future trajectories for objects in the environment, even when partially or completely occluded, and take appropriate action to avoid potential collisions.

In the literature, one focus of occlusion-aware methods is to provide provable safety in scenarios with occlusions. Reachability analysis can be used to represent the set of possible states that a potentially occluded vehicle can reach and to plan safe trajectories. Orzechowski~\etal~\cite{Orzechowski_2018} propose an approach to predict the presence of a vehicle coming out of an occluded region using reachable sets.  These reachable sets cover all possible future states of the possible hidden vehicle under a set of assumptions regarding their initial velocity, position, and traffic rules. If the potential trajectory of the EV does not intersect with any of the reachable sets of the other obstacles, it is verified as safe. If the verification fails, the intended trajectory will be iteratively adapted until a verifiably safe trajectory is found. Neel and Saripalli \cite{Neel_2020} reduce the bounds on the state intervals used for predicting the future occupancy of hidden vehicles by tracking the occlusions. By doing so, they restrict the position interval and the upper limit on the velocity of a possible hidden vehicle, producing smaller predicted sets. Sánchez~\etal~\cite{Sánchez_2022} use sequential reasoning based on reachability analysis and previous observations to reduce the occluded areas and remove infeasible obstacle states. Although the safety of these methods is demonstrated, they can lead to conservative driving behavior in some situations.


Another focus of researchers is on estimating collision risk caused by the potential hidden vehicles in occlusions. Yu \etal~\cite{Yu_2019} proposes an algorithm that quantifies the risk caused by limited sensing capabilities and geometric occlusions, leveraging the known road layout to quantify risk in the environment. It applies a sampling method to represent potentially occluded agent states as particles, and forward propagate each particle assuming constant velocity. These particles are then used to calculate the risk of collision. Wang \etal~\cite{Wang_2022} proposes the use of a Dynamic Bayesian Network to infer and quantify the potential risk at visually occluded areas. 

An additional frequently used approach to handle occlusions is Partially Observable Markov Decision Processes (POMDP). Hubmann \etal~\cite{Hubmann_2019} propose a behavior planner based on a POMDP formulation that explicitly considers possibly occluded vehicles, where the future FOV of the autonomous car is predicted over the whole planning horizon. Using Monte Carlo sampling they generate possible future episodes, including the possible hidden vehicles, that are used to derive an optimized policy for the autonomous vehicle. Zhang \etal ~\cite{Zhang_2021} infers the appearance probability of phantom objects along with their future movement using map information and road topology. These context-aware phantom road users are incorporated within the POMDP-based behavior planner, which is solved online by constructing a Monte Carlo tree with reachable state analysis. Although POMDP solutions can avoid too conservative strategies, they cannot guarantee safety \cite{Wang_2023}.

Lately, many learning-based approaches have tried to tackle the problem of reduced visibility. Bouton~\etal~\cite{Bouton_2019} presented a safe reinforcement learning algorithm based on a model-checker to ensure safety guarantees with a belief update technique to make the decision strategy robust to perception errors and occlusions. Kamran \etal~\cite{Kamran_2020} use a risk-aware DQN to deal with occluded intersections. It includes a risk-based reward function, which punishes risky situations as well as collision failures. The main disadvantages of these approaches are their sensitivity to the hyperparameters and their lack of interpretability. 

As a conclusion, the safety of autonomous vehicles is significantly compromised by limited visibility caused by obstructed views or sensor range constraints. However, the methods outlined may yield outcomes that are either overly cautious, unsafe, or challenging to interpret. In this paper, the Multimodal Interaction-Aware Motion Prediction (MultIAMP) framework (\cite{vinicius_ACCESS, vinicius_TIV}) is adapted in order to take into account the occlusions and generate more reliable predictions in situations where the FOV of the EV is obstructed. It is done by creating hidden vehicles and virtual corridors from the occluded zones, that are classified and tracked over time through sequential reasoning. The occlusion-aware MultIAMP is evaluated in two use cases: an overtaken and a left turn at an intersection, using a state-of-the-art motion planner.

The outline of this paper is organized as follows: Section \ref{sec: multiamp_framework} summarizes the main features of the MultIAMP framework. Section~\ref{sec: framework} introduces the designed layer to tackle occlusions. Section~\ref{sec: experimental_results} presents the evaluation of the extended framework in two driving scenarios. Finally, Section~\ref{sec: conclusion} provides some concluding remarks.

\section{Multimodal Interaction-Aware Motion Prediction}
\label{sec: multiamp_framework}

The MultIAMP framework uses a combination of Dynamic Bayesian Network (DBN) and Markov chains to infer the intentions and predict the motion of the surrounding vehicles. It explicitly considers interactions, road layouts, and traffic rules.

At each time step, the vehicle-to-vehicle and vehicle-to-layout interactions are taken into account to infer the probability of stopping or crossing the intersections, the probability of being in each of the possible navigable corridors, as well as the probability of changing lanes. These intentions are fused with the motion predictions computed with a kinematic model to result in a motion grid $\mathcal{M}$. This grid comprises probability distributions that depict potential trajectories for the surrounding vehicles and is used by the EV to navigate through the scene.

This probabilistic approach is a balanced trade-off between point-based and set-based motion prediction methods. While the former can be considered under-conservative due to its lack of consideration for prediction errors, and the latter is regarded as over-conservative as it considers every possible motion of the vehicle, the implemented approach combines the advantages of both methodologies.

The grid-based motion prediction allows for a more comprehensive understanding of the environment by considering multiple possible outcomes. This probabilistic approach enables the EV to make informed decisions based on the likelihood of different scenarios, enhancing its ability to navigate complex and dynamic situations.

The framework can be divided into four main blocks: \textit{Corridors}, \textit{Relations}, \textit{Intentions}, and \textit{Motion Prediction}. Each block will be briefly described below. For more details, the reader is referred to \cite{vinicius_ACCESS, vinicius_TIV}.

\begin{itemize}
    \item Corridors: is a goal-oriented search where each corridor $\zeta$ represents a potential path a vehicle might follow. The extension of these corridors is limited to the distance that the vehicle can travel in a given time interval at its current velocity, assuming constant acceleration. 
    
    \item Relations: the relations obtained from the list of corridors and the map are threefold: lateral relations, corridor-to-intersection (distance to intersections), and corridor-to-corridor (corridors' dependencies).
    
    \item Intention estimation: a DBN is used to infer the intentions of the surrounding vehicles at the scene. The DBN computes the expected and intended longitudinal and lateral maneuvers of each vehicle and infers the intentions using a particle filter. It takes into account the corridors and relations and outputs the probability of being in each corridor as well as its longitudinal probability of stopping at the next intersection. 
    
    \item Motion prediction: the motion prediction of the surrounding vehicles is computed based on a Markov chain abstraction of a motion model. It takes as input the corridors, their relations with each other, and the probabilities inferred with the DBN. With these inputs, acceleration profiles are obtained and applied to the computation of the longitudinal motion prediction of each corridor. These predictions are projected onto their corridors and combined to form the motion grid $\mathcal{M}$.
\end{itemize}

\section{Framework adaptation}
\label{sec: framework}

\begin{figure}
    \centering
    \includegraphics[width=0.96\linewidth]{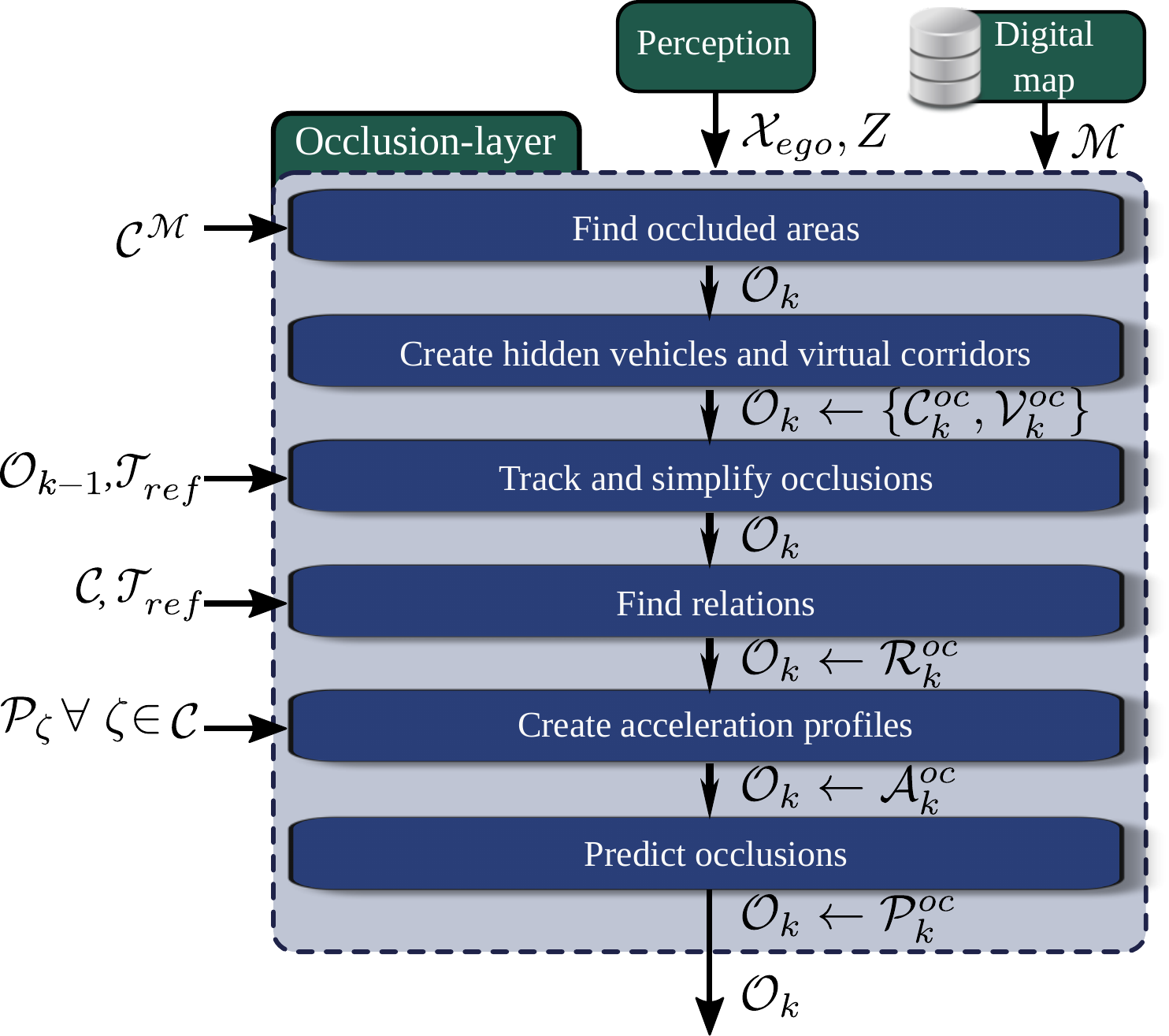}
    \caption{Flowchart of the occlusion layer in MultIAMP.}
    \label{fig: flowchart}
\end{figure}

To handle occluded areas within the MultIAMP framework, it was necessary to consider some additional steps to search and track the occlusions, find their relations among each other and with the other visible vehicles, and generate their motion predictions. These steps are presented in the flowchart in Fig.~\ref{fig: flowchart} and described below. It takes as input the current state of the EV ($\mathcal{X}_{ego}$) and of the SVs ($Z$) and a digital map $\mathscr{M}$.

\subsection{Search and classify}

\begin{figure}
    \centering
    \includegraphics[width=0.94\linewidth]{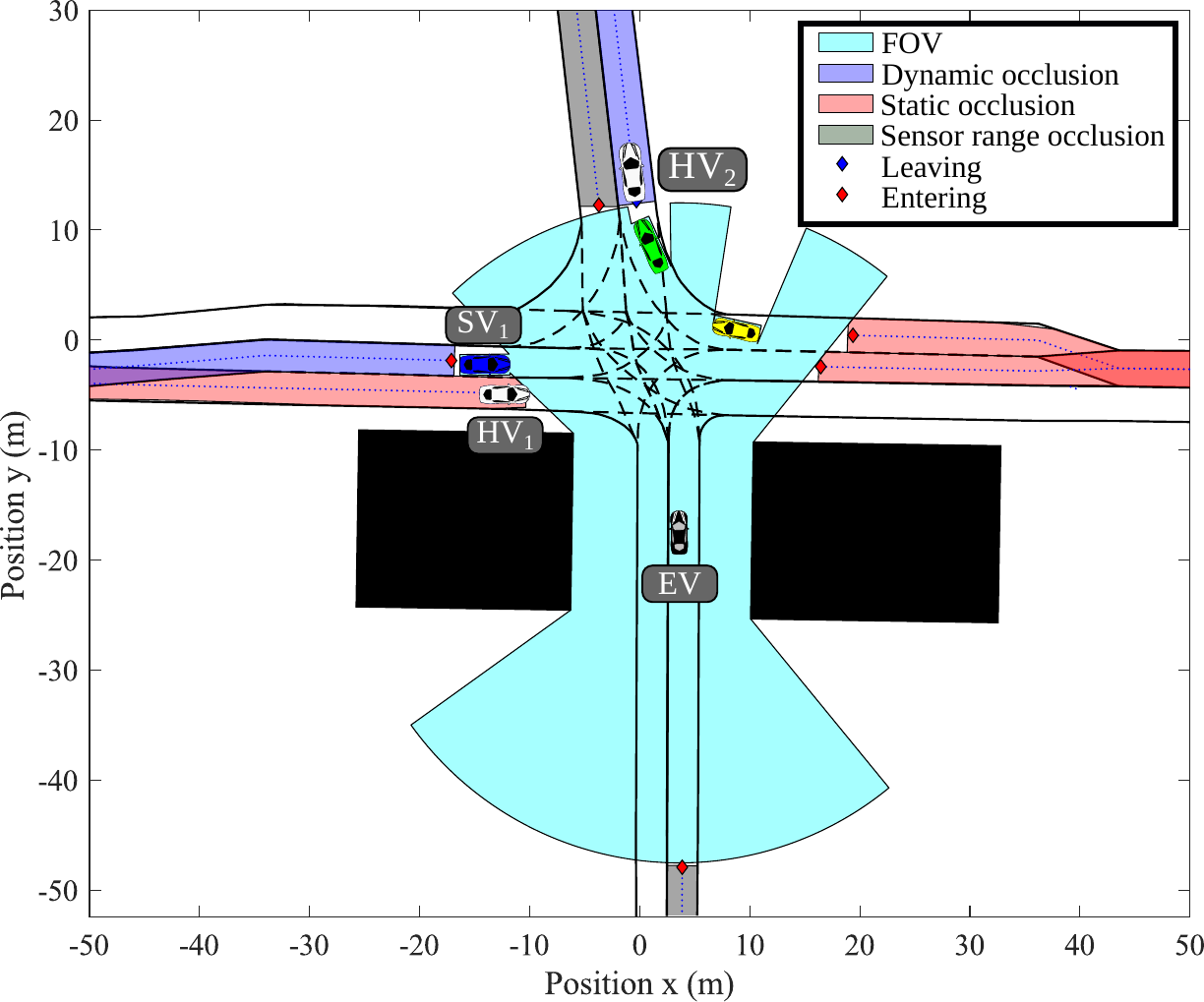}
    \caption{Example of an occluded situation.}
    \label{fig: occlusions_FOV}
    \vspace{-15pt}

\end{figure}
The FOV of the EV (e.g. Fig.~\ref{fig: occlusions_FOV}) is represented by a polyshape centered at the EV's location, with the maximum distance between the EV's position and the edges of the polyshape being determined by the maximum sensor range. Besides the reached distances, it also stores the information of the entities that were hit. Notice that vehicles that are only partially visible are also considered as being fully viewed.

To determine the occluded zones $\mathcal{O}_k$ at time step $k$, all corridors $\mathcal{C}^{\mathscr{M}}$ within the limited section of the map where the EV is driving are found. This corridor search is performed during the first iteration of the algorithm, and the resulting corridors are stored and used until a new search becomes necessary (when they are no longer a good representation of the surroundings). Furthermore, the polygons of these corridors are intersected with the EV's FOV, and the areas that do not intersect, if bigger than a threshold, are considered as occlusions (see the blue, red, and gray areas in Fig.~\ref{fig: occlusions_FOV}).

It is assumed that a vehicle can only emerge from within the predictions of the previous time step. To implement this, the current occlusion area is intersected with the occlusions of the previous step. Moreover, in order to avoid missing some occluded areas, vehicles that were previously seen and are not present in the current time step are kept alive by including their prediction of the first time step of the prediction horizon in the resulting occluded zone.

\begin{figure*}[ht!]
    \centering
    \subfloat[]{\includegraphics[width=0.48\linewidth]{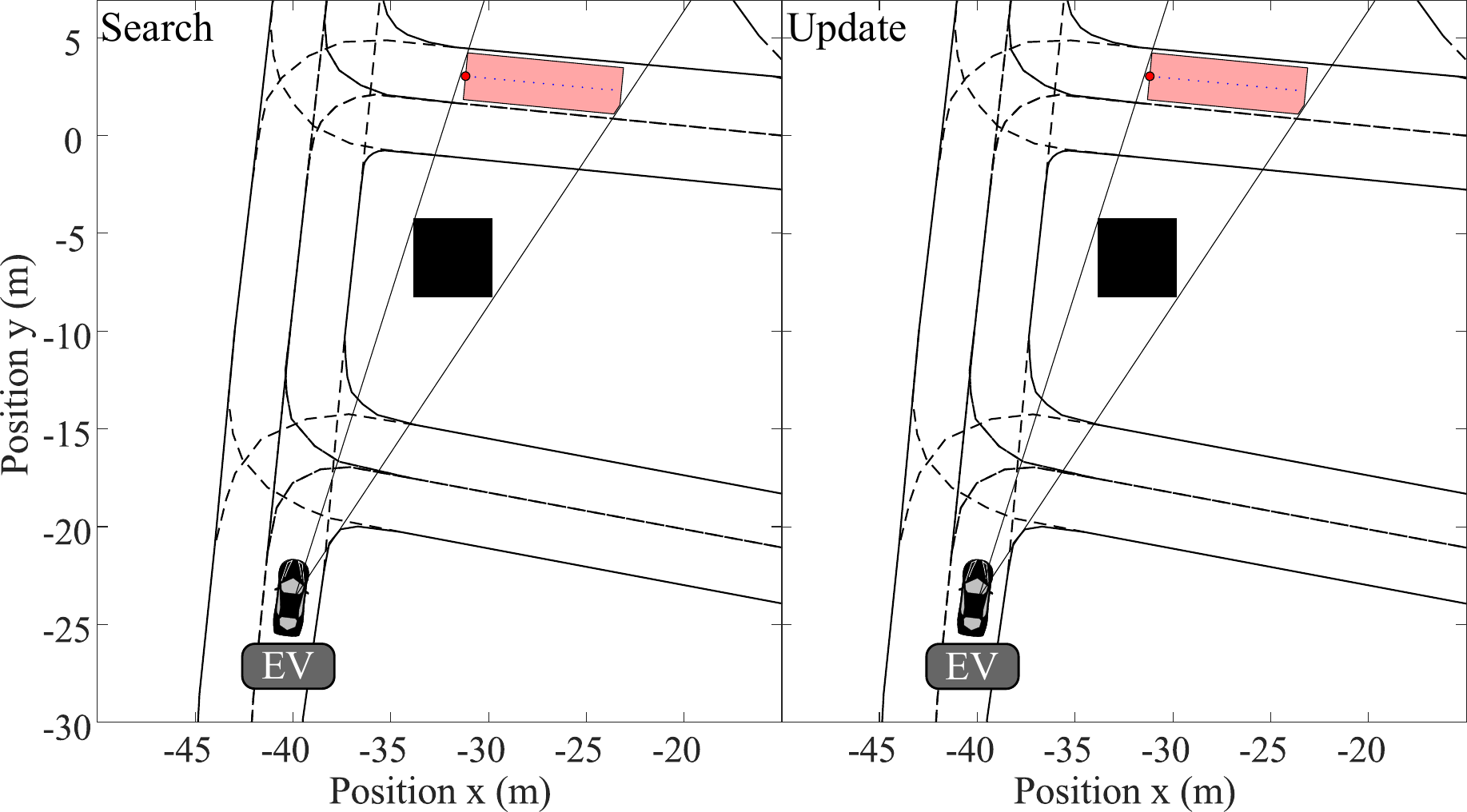}\label{fig: update_position_occlusion_a}} 
    \subfloat[]{\includegraphics[width=0.48\linewidth]{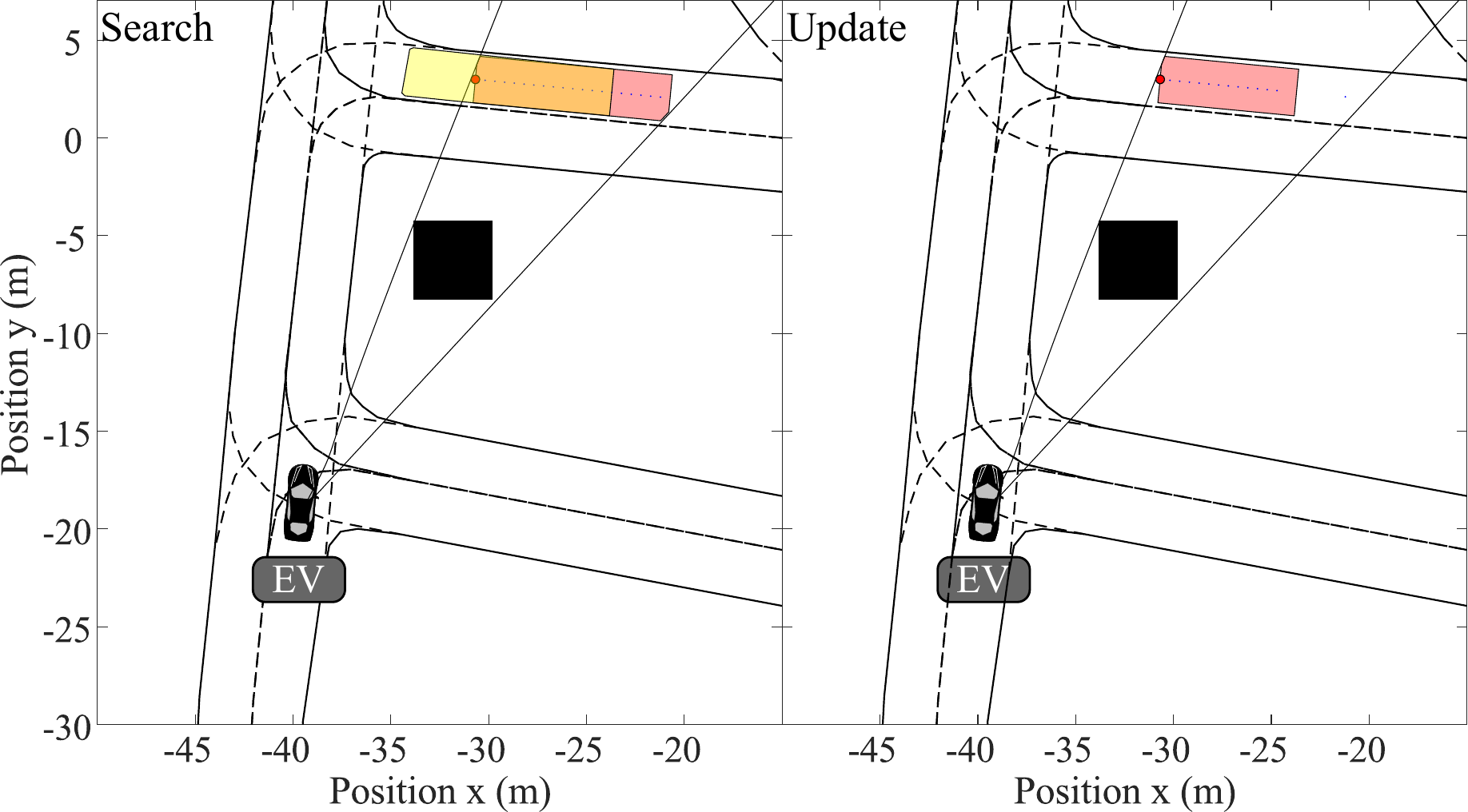}\label{fig: update_position_occlusion_b}}\\
    \subfloat[]{\includegraphics[width=0.48\linewidth]{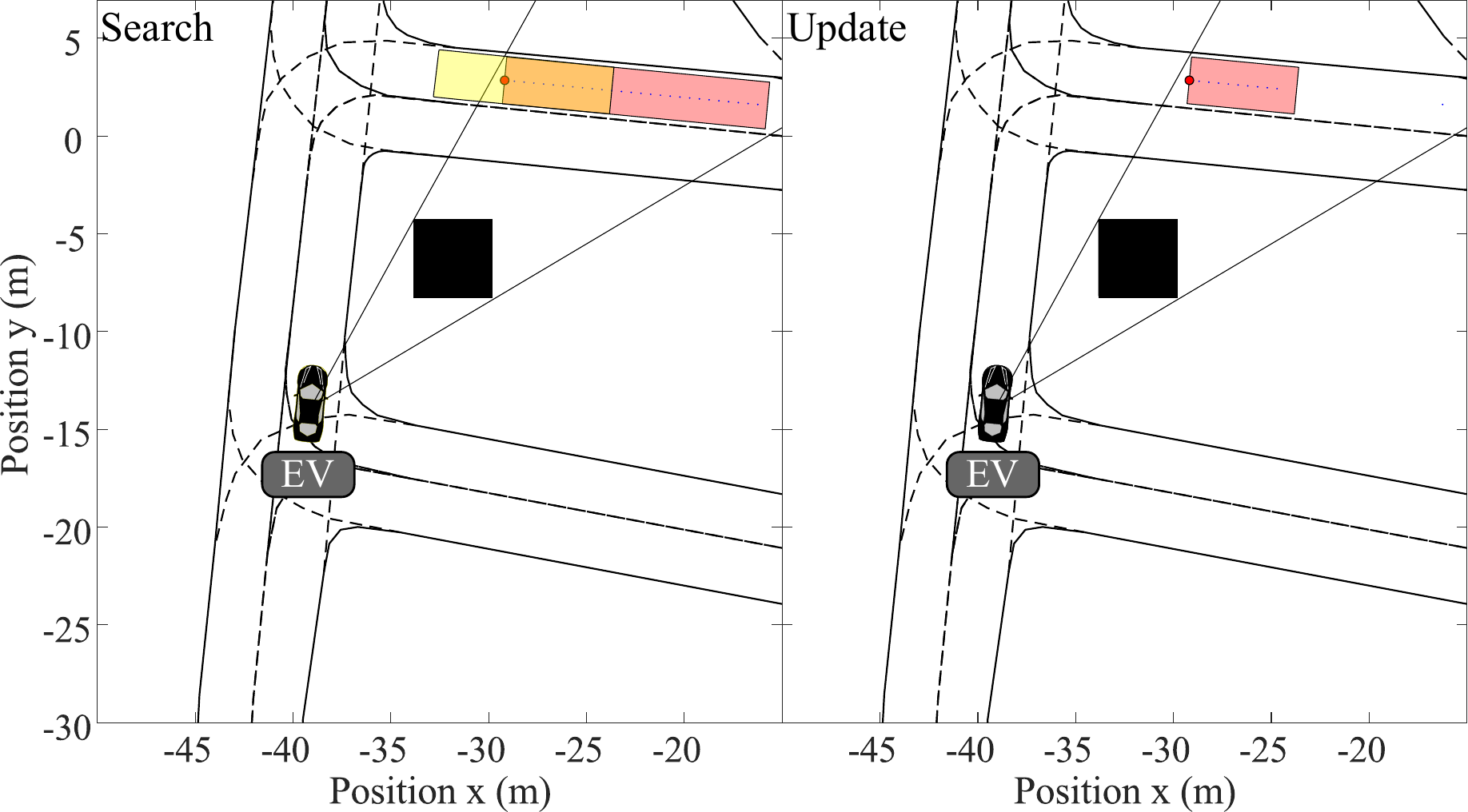}\label{fig: update_position_occlusion_c}} 
    \subfloat[]{\includegraphics[width=0.48\linewidth]{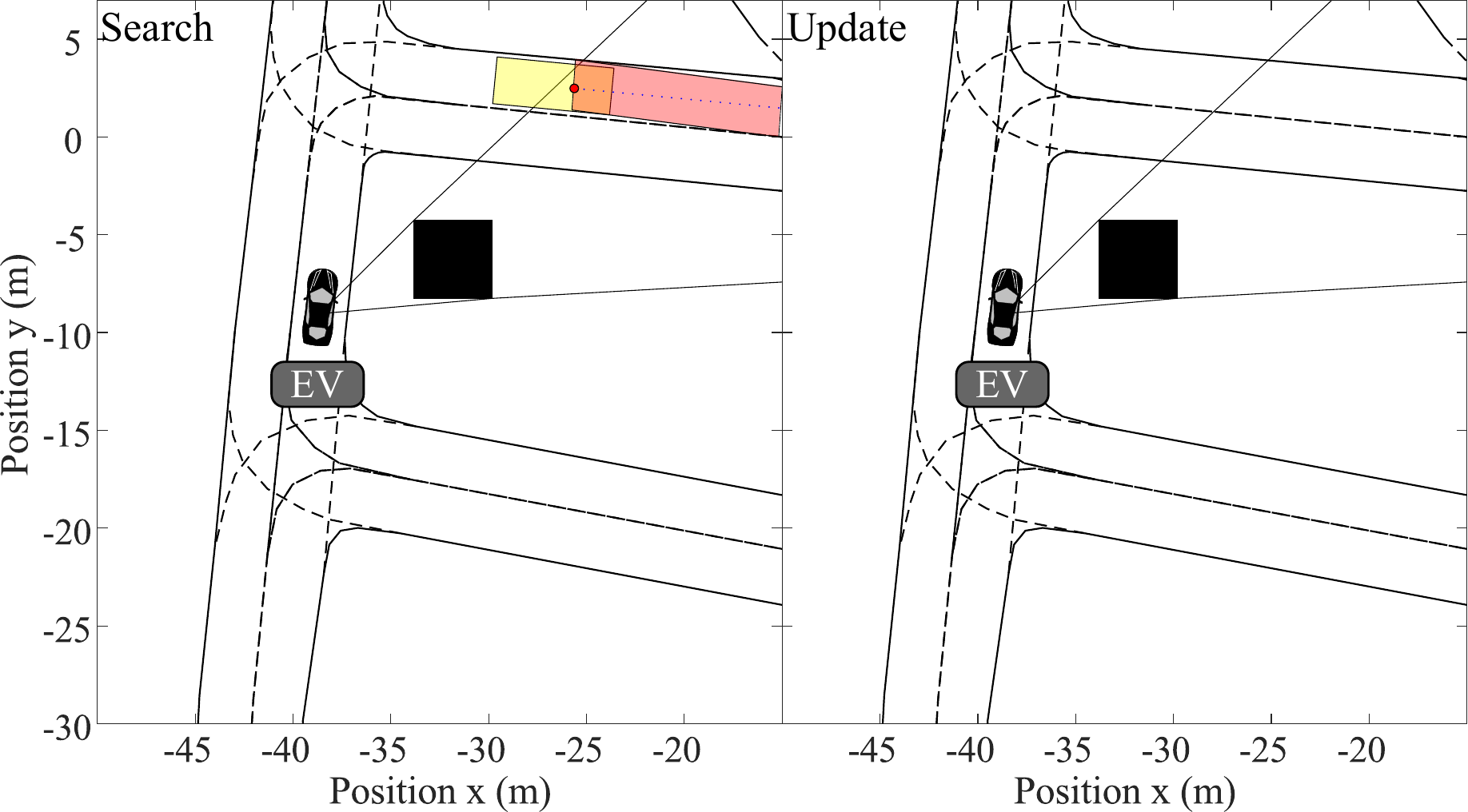}\label{fig: update_position_occlusion_d}}
    \caption{Evolution of an occlusion at (a) 0 s, (b) 1.0 s, (c) 2.0 s and (d) 3.0 s. \textit{Search} represents (in red) the occlusion found in the current time step and \textit{update} shows the occlusion intersected with the prediction of the previous time step (in yellow). The intersection between the two areas is displayed in orange.}
    \label{fig: update_position_occlusion}
    \vspace{-5pt}

\end{figure*}

The effect of the intersection with previous occlusions and prediction areas can be seen in Fig.~\ref{fig: update_position_occlusion}. In this example, the EV is traveling through a main lane approaching an intersection. Due to a static obstacle, an occlusion is formed in an arm of the intersection that has to yield. The situation begins in Fig.~\ref{fig: update_position_occlusion_a}, where the searched and updated occlusions are the same. As the EV advances in its path, the initial occluded area gets smaller due to its intersection with the prediction of the previous time step (in yellow) until the point where it disappears in Fig.~\ref{fig: update_position_occlusion_d}.

With the information from the FOV, the occlusions can be categorized as being caused by static obstacles (red), dynamic obstacles (blue), or sensor range (gray) \cite{thesis_Victor}, as can be seen in Fig.~\ref{fig: occlusions_FOV}. Furthermore, they are also categorized according to the way they interact with the EV's FOV: they can be either entering (red rhombuses) or leaving (blue rhombuses) the FOV. Additionally, occlusions are classified as either \textit{bounded}, when both the occlusion's start and end are visible to the EV, or \textit{unbounded} when only one of the bounds is visible.

Since there are overlaps between the corridors that originate the occlusions, the occlusions are intersected with each other, and their originating corridors are compared. If they intersect and share some lanelets, they are joined. For instance, in Fig.~\ref{fig: occlusions_FOV} the gray occlusion behind the EV can be originated from three corridors from $\mathcal{C}^{\mathscr{M}}$ (one turning left, one going straight, and one turning right). Since these corridors share the lanelet that contains the resulting occlusion, they are joined and only one occlusion is created.

\subsection{Hidden vehicles and virtual corridors}

At the beginning or end of the occlusion, depending on whether it is entering or leaving the FOV, a hidden vehicle (HV $\in \mathcal{V}^{oc}_{k}$) is placed. For clarity, in Fig.~\ref{fig: occlusions_FOV} only two HVs are displayed ($HV_1$ that is entering and $HV_2$ that is leaving the FOV), but one is created for every rhombus that appears. 

For each HV, the possible virtual corridors $\mathcal{C}^{oc}_{k}$ are determined via a graph search where each corridor $\zeta^{oc}$ represents a potential path the HV might follow. The corridors' maximum length is determined by the maximum allowable velocity.

\subsection{Track and simplify occlusions}

In order to gain some information about the vehicles that might be occluded, the occlusions are tracked. The approach implemented is inspired by the ones from~\cite{Neel_2020, Wang_2021}, where the occlusion's velocity limits are updated at every time step. Here, probabilistic distributions are used to represent the velocity and are updated when new information is received.

The occlusions are tracked based on their area and originating corridors $\zeta^{\mathscr{M}} \in \mathcal{C}^{\mathscr{M}}$. To achieve this, each new occlusion is intersected with the previous occlusions $\mathcal{O}_{k-1}$, and their originating corridors are compared. If the resulting intersected area is bigger than a threshold and if they share the same lanelet, both occlusions are considered the same. When a new occlusion represents the updated version of an occlusion from the previous time step, its initial probability distribution $p(t_k)$ can be revised, eliminating the necessity to encompass all velocity intervals. $p(t_k)$ contains the probability of each cell that composes the state space of the motion prediction model, obtained from a bivariate distribution (longitudinal position and velocity). At first, the marginal velocity distribution spans the entire range, from 0 m/s up to the maximum allowed velocity (15 m/s in this instance). For the longitudinal position distribution, its range is limited to either the bounds of the occlusion or by a maximum distance. Notice that only bounded\footnote{Occlusions that have a well-defined area.} occlusions can undergo an update to their initial distribution.

\begin{figure}[ht]
    \centering
    \subfloat[]{\includegraphics[width=0.75\linewidth]{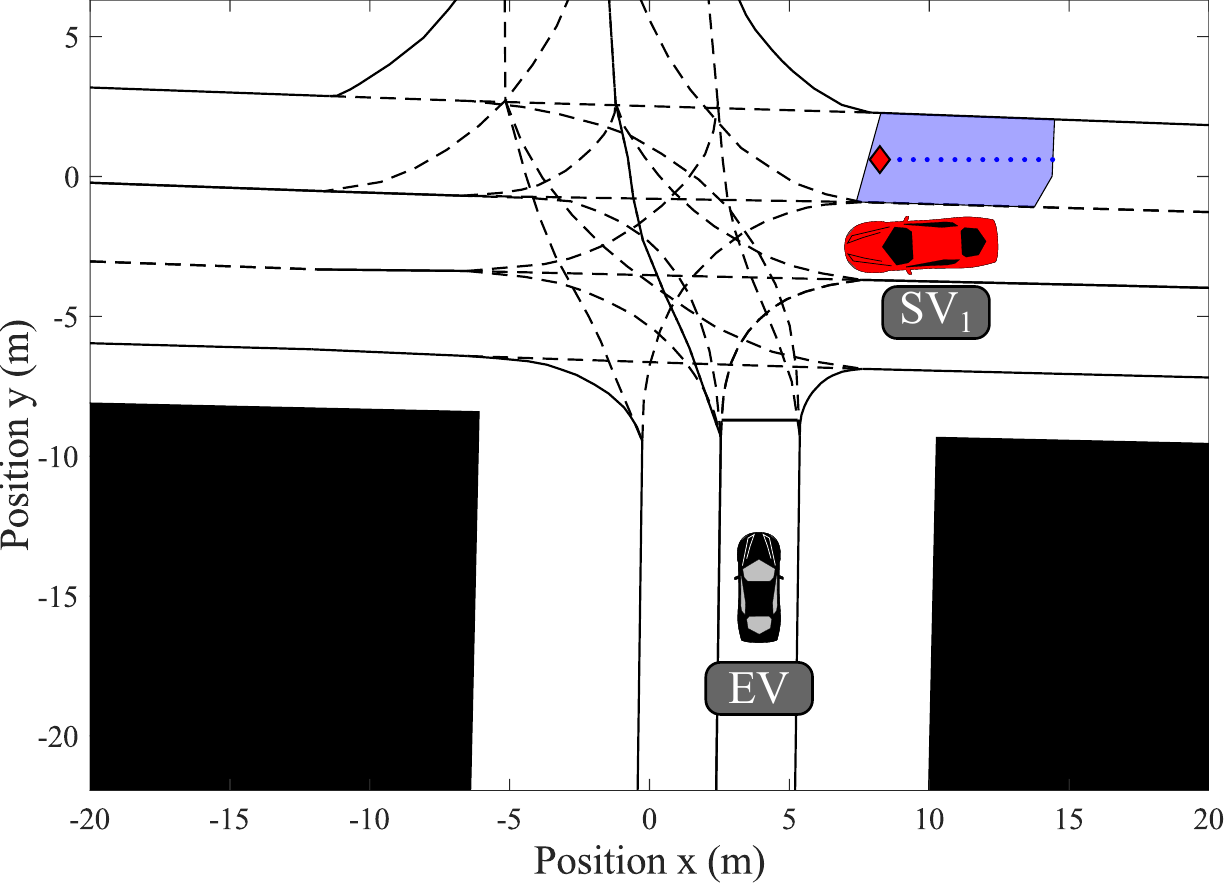}\label{fig: updateVel_occlusionsA}} \\
    \subfloat[]{\includegraphics[width=0.75\linewidth]{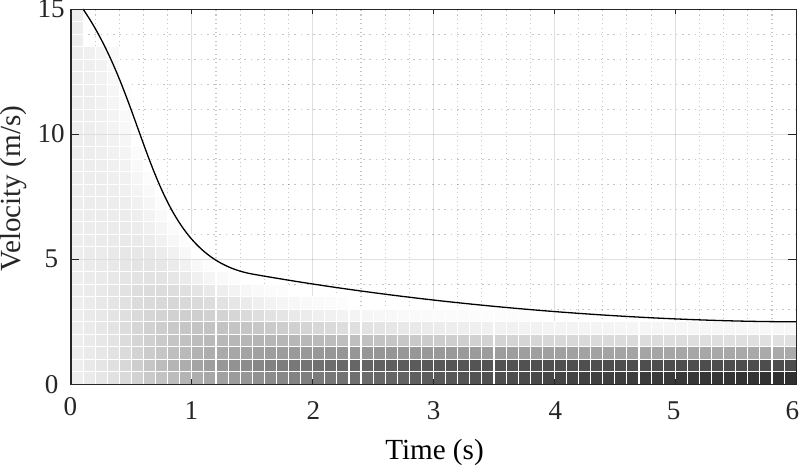}\label{fig: updateVel_occlusionsB}}
    \caption{Example of a (a) bounded occlusion and (b) the evolution of the initial velocity distribution.}
    \label{fig: updateVel_occlusions} 
    \vspace{-10pt}

\end{figure}

Fig.~\ref{fig: updateVel_occlusionsA} shows the initial position of a situation that will be used to illustrate the process of updating the velocity limits in the initial probability distribution $p(t_k)$. The EV is approaching an intersection where vehicle $SV_1$ is waiting to turn left. Behind $SV_1$ there is a dynamic occlusion caused by $SV_1$. To simplify the example, only this occlusion is represented.

Initially, the framework has no information regarding what could be in the occluded area (notice that the velocity axis reaches the maximum allowed velocity in $t=0~s$ in Fig.~\ref{fig: updateVel_occlusionsB}). As time progresses, the algorithm starts to limit the velocity of what can be within the occlusion bounds, since a vehicle with a velocity bigger than this limit would have appeared in this time step. 

To do so, the initial probability distribution $p(t_k)$ is extracted from the first prediction of the occlusion from the previous time step that corresponds to the current step. This extraction involves intersecting the occlusion's area with the prediction. Subsequently, the intersected values of the distribution within the prediction are obtained and normalized.

Certain occlusions may not interact with the EV's trajectory $\mathscr{T}_{ref}$. In order to simplify the number of occlusions and avoid unnecessary computation, any corridor that does not intersect with the EV's trajectory, or has a closer dependency caused by another occlusion, is removed. This approach aims to simplify the analysis by focusing only on relevant occlusions.

\subsection{Find relations between corridors and layout}

The relations between corridors and layout $\mathcal{R}^{oc}_{k}$ are the output of the block "Find relations". They are necessary in order to determine which real vehicle and/or scenario component will influence the prediction from the hidden vehicle.

The virtual corridors are intersected with the corridors from visible vehicles $\mathcal{C}$ and with the trajectory $ \mathscr{T}_{ref}$ of the EV in order to find their dependencies (which corridor will influence their predictions). To do so, the centerlines are pairwise intersected, and the closest and longest dependency is selected to constrain the predictions of the occlusion \cite{vinicius_TIV}.

In scenarios where the virtual corridor has to yield, the intention of the corridor will be to stop at the yield line. The distance to the intersection is computed by finding the index of the intersection's entrance on the corridor's centerline.

Notice that the dependencies are only searched for virtual corridors. The other vehicles are assumed to have complete awareness, and therefore, occlusions are not taken into account as constraints on their motion.

\subsection{Acceleration profile}

The prediction framework in MultIAMP takes as input acceleration profiles that try to represent the nuances of the vehicle's motion within the prediction horizon (acceleration, braking, constant velocity). For this reason, for each corridor $\zeta$ from a HV, an acceleration profile $\alpha_{\zeta}^{\text{hv}} \in \mathcal{A}_{k}^{oc}$ is created. It takes into consideration the interactions with a leading vehicle (using its predictions $\mathcal{P}_{\zeta}$) and the interactions with the layout.

The profiles obtained for real vehicles in MultIAMP (described in~\cite{vinicius_TIV}) initiate with the vehicle's current velocity and acceleration. This information is readily available for visible vehicles. However, for occluded vehicles, it is necessary to estimate their velocity. To achieve this, each profile is computed three times, varying the initial velocity within the allowed limits. In contrast to profiles for visible vehicles, where three distinct profiles are derived considering different restrictions (curvature, obstacle, or intersection), the profiles computed for occluded vehicles encompass all restrictions simultaneously. This is based on the assumption that the potential HV will adhere to traffic rules and the presence of other vehicles in its surroundings. The resulting profile is the weighted average of these three computations, where the weights are empirically defined.

\subsection{Generate predictions}

The predictions $\mathcal{P}_{k}^{oc}$ of occlusions are computed similarly to the predictions of visible vehicles. The main difference is that, since no information about its pose is given, the initial distribution covers the area of the whole occlusion and the space between the velocity limits. In cases where the occlusion is confined within boundaries, both area and velocity ranges can be restricted, as shown in the examples depicted in Fig.~\ref{fig: update_position_occlusion} and Fig.~\ref{fig: updateVel_occlusions}, respectively. However, when the occlusion is unbounded, the area is limited to remain closer to the FOV, while the velocity must cover the entire interval due to the absence of any available information.
\begin{figure*}[!ht]
    \centering
    \includegraphics[width=0.75\linewidth]{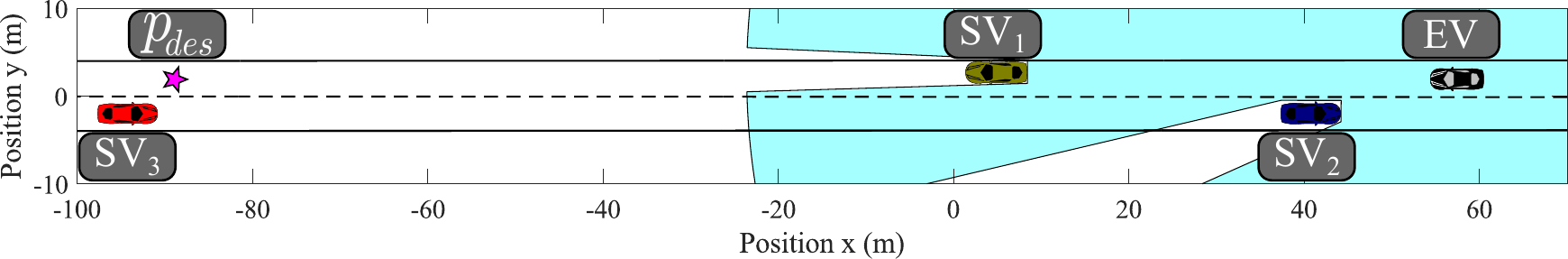}
    \caption{Highway scenario layout. The area in cyan is the EV's FOV. Vehicles outside the FOV are not perceived by the EV.}
    \label{fig: initialPos_occlusions_highway}
\end{figure*}

\section{Experiments}
\label{sec: experimental_results}

The main goal of the prediction module in the vehicle architecture is to provide accurate predictions from the surrounding vehicles to a motion planner module, so that the latter can generate trajectories that make the EV safely drive through traffic. The incorporation of occlusion handling in the framework adds another layer of protection to the system, making the motion planner produce more cautious trajectories in situations where, without this module, the trajectories could be unsafe.

In this context, the occlusion-aware MultIAMP is evaluated in two scenarios where occlusions are presented, considering its influence in the motion planner stage. For this assessment, a state-of-the-art motion planner \cite{Juan_IV2} will be considered. Since the main goal is to ensure safe planning without compromising comfort and efficiency, the performance of the experiments is measured with different planning-oriented metrics. It is important to mention that the configuration parameters of the motion planner are kept the same for all experiments.

The selected motion planning algorithm creates a set of possible trajectories and selects the best of them according to a merit function that combines four decision variables, described below. Each trajectory candidate is formed by a curve and a speed profile. The curve is generated using quintic Bézier curves primitives, and the speed profile is computed by taking into account comfort requirements, traffic signals and the reachable sets of the SVs. Each candidate is evaluated using a set of Trajectory Performance Indicators that consider different data, like maximum accelerations or jerks, distance to obstacles, lane invasion, and average speed, among others. Each trajectory is evaluated using a set of four Decision Variables (DV): longitudinal comfort, lateral comfort, safety and utility, which are weighted and combined using a merit function described in \cite{juan_sensors_2021}.

The interaction between the key modules of the ADS happens as follows: the motion prediction module provides the motion planner with the motion grid $\mathcal{M}$ containing the predicted occlusions and receives in return the intended trajectory for the EV. In experiments involving occlusions, the interaction is unidirectional. The EV reacts to the predictions from the surrounding visible vehicles and occlusions. However, since it does not have the right-of-way over others, its intended motion will not influence other vehicles. This is due to their adherence to a constant velocity model, neglecting inter-vehicle interactions.

Every situation is simulated three times:

\begin{itemize}
    \item omniscient: all vehicles are always visible;
    \item occlusion-unaware: the FOV is limited, but the framework does not generate predictions for the occlusions;
    \item occlusion-aware: the FOV is limited, and the framework generates predictions for the occlusions.
\end{itemize}

The Key Performance Indicators (KPIs) used to evaluate the behavior of the motion planner considering the different types of predictions are summarized in Table~\ref{tab: kpi_definition}. The longitudinal and lateral acceleration KPIs evaluate the maximum value of these variables during the experiment; the longitudinal and lateral jerk are computed from the average value of the jerk. These four metrics are used to measure the comfort of the experiment. The risk-to-collision indicator is computed using the Risk from the Time to Closest Encounter (RTTCE) metric proposed in \cite{eggert_2017}, by comparing the position of the EV with the SVs all along the experiment. Finally, the travel time quantifies the time it took for the EV to reach the goal position.

\begin{table}
\centering
\caption{Performance Indicators \cite{vinicius_TIV2}.}
\label{tab: kpi_definition}
\setlength{\tabcolsep}{4mm}
\setlength\extrarowheight{4pt}
    \begin{tabular}{|l|l|}
        \hline
        \bf{Parameter} & \bf{Formula} \\[2pt]
        \hline
        Longitudinal Acceleration
                & $\max(|\gamma_x(s)|)$ 
                \\[2pt]
        \hline
        Longitudinal Jerk
                & $\overline{|\jmath_x(s)|}$
                \\[2pt]
        \hline
        Lateral Acceleration
                & $\max(|\gamma_y(s)|)$
                \\[2pt]
        \hline
        Lateral Jerk
                & $\overline{|\jmath_y(s)|}$
                \\[2pt]
        \hline
        Risk-to-collision
                & $\overline{RTTCE} + \max(RTTCE)$
                \\[2pt]
        \hline
        Travel time
                & $t_N$
                \\[2pt]
        \hline
        \end{tabular}
\end{table}

\subsection{Highway scenario}

In the first scenario, the EV is traveling through a two-lane road where the invasion of the opposite lane is allowed for overtaking (see Fig.~\ref{fig: initialPos_occlusions_highway}). In its path, there is a vehicle stopped $SV_1$ in the center of the lane. In the opposite direction, vehicle $SV_2$ is coming. After $SV_2$ passes the EV's position, the EV is faced with the decision to either overtake $SV_1$ or wait until it starts moving. The destination point $(p_{des})$ is $145~m$ away from the initial position of the EV.

\begin{figure}[ht!]
    \centering
    \subfloat[]{\includegraphics[width=\linewidth]{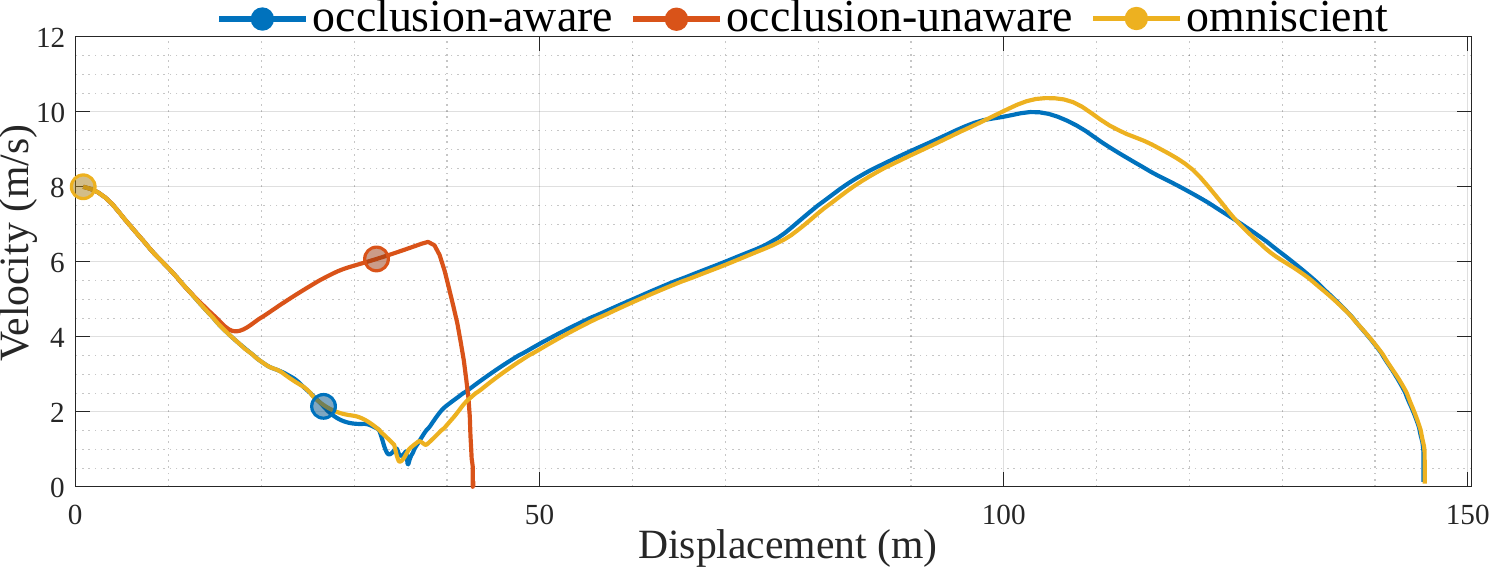}%
    \label{fig: vel_highway}}
    
    \subfloat[]{\includegraphics[width=\linewidth]{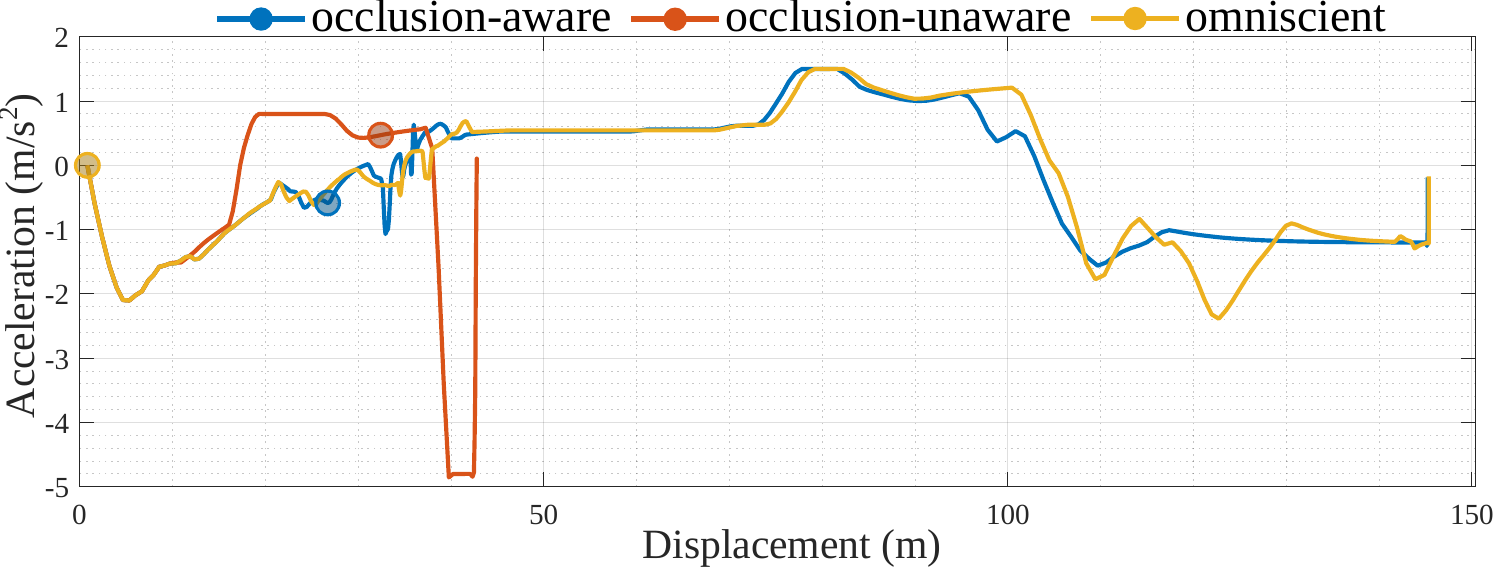}%
    \label{fig: acc_highway}}
    \caption{Velocity and acceleration profiles for the different experiments on the highway scenario. (a) Velocity profile. (b) Longitudinal acceleration profile.}
    \label{fig: motion_highway}
\end{figure}

Fig.~\ref{fig: motion_highway} presents the velocity and acceleration over the path for the three experiments. The marked points represent the moment the EV perceives $SV_3$.

In the occlusion-unaware experiment, the EV plans to do a lane change once $SV_2$ has passed. It is a reasonable choice since $SV_1$ is stopped and for the EV, nothing comes in the other direction (as can be seen by the FOV represented in cyan in Fig.~\ref{fig: initialPos_occlusions_highway}). 

At $4.4~s$ after the planned lane change, the EV perceives $SV_3$ (marked point in Fig.~\ref{fig: motion_highway}), but it has already invaded the adjacent lane and is not able to find a viable candidate that can lead it to a safe position. Hence, it comes to a full stop in the middle of the lane. The experiment is stopped at the moment the EV collides with $SV_3$.

In the occlusion-aware experiment, the EV receives the predictions (included in the motion grid $\mathcal{M}$) from a HV coming in the opposite direction. These predictions are converted into possible collision points \cite{Juan_IV1} that prevent the lane change. Notice that for the EV there is no distinction between visible vehicles and possible HVs. The EV almost comes to a full stop behind $SV_1$ (at displacement $s \approx 35~m$) and, once $SV_1$ starts to move, it continues its journey towards the destination point. Similar results are achieved with the omniscient version.

The differences in velocity and acceleration between the occlusion-aware and omniscient versions after displacement $s \approx 90~m$ arise from the fact that the occlusion-aware sends to the EV the information that exists a possible HV in the same lane as the EV at the end of the FOV traveling in the same direction. In contrast, in the omniscient version, this does not happen since it has full awareness of the scene. This potential HV affects the planned velocity of the EV, causing a reduction in its maximum attainable value.

\begin{figure}[!ht]
    \centering
    \includegraphics[width=0.9\linewidth]{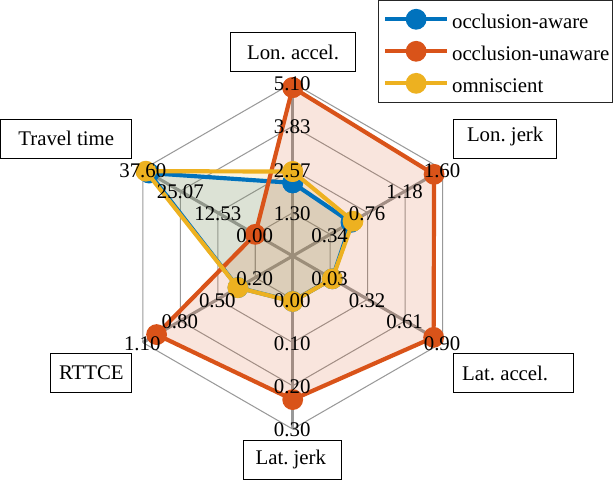}
    \caption{KPIs for the highway experiment.}
    \label{fig: radar_occlusion_highway}
\end{figure}

Fig.~\ref{fig: radar_occlusion_highway} displays the performance indicators of the three experiments. The definition of these KPIs requests a smaller-is-better (SiB) format. Hence, the closer to the center of the figure, the better the performance of the KPI. The occlusion-unaware experiment was unable to reach the goal due to the collision with $SV_3$, which is represented as 0 in the travel time of Fig.~\ref{fig: radar_occlusion_highway}. Apart from the risk due to the collision, the EV faced higher values of acceleration and jerk in the occlusion-unaware experiment, resulting also in lower comfort levels.

\vspace{-5pt}

\subsection{Intersection scenario}

\begin{figure}[!ht]
    \centering
    \includegraphics[width=0.9\linewidth]{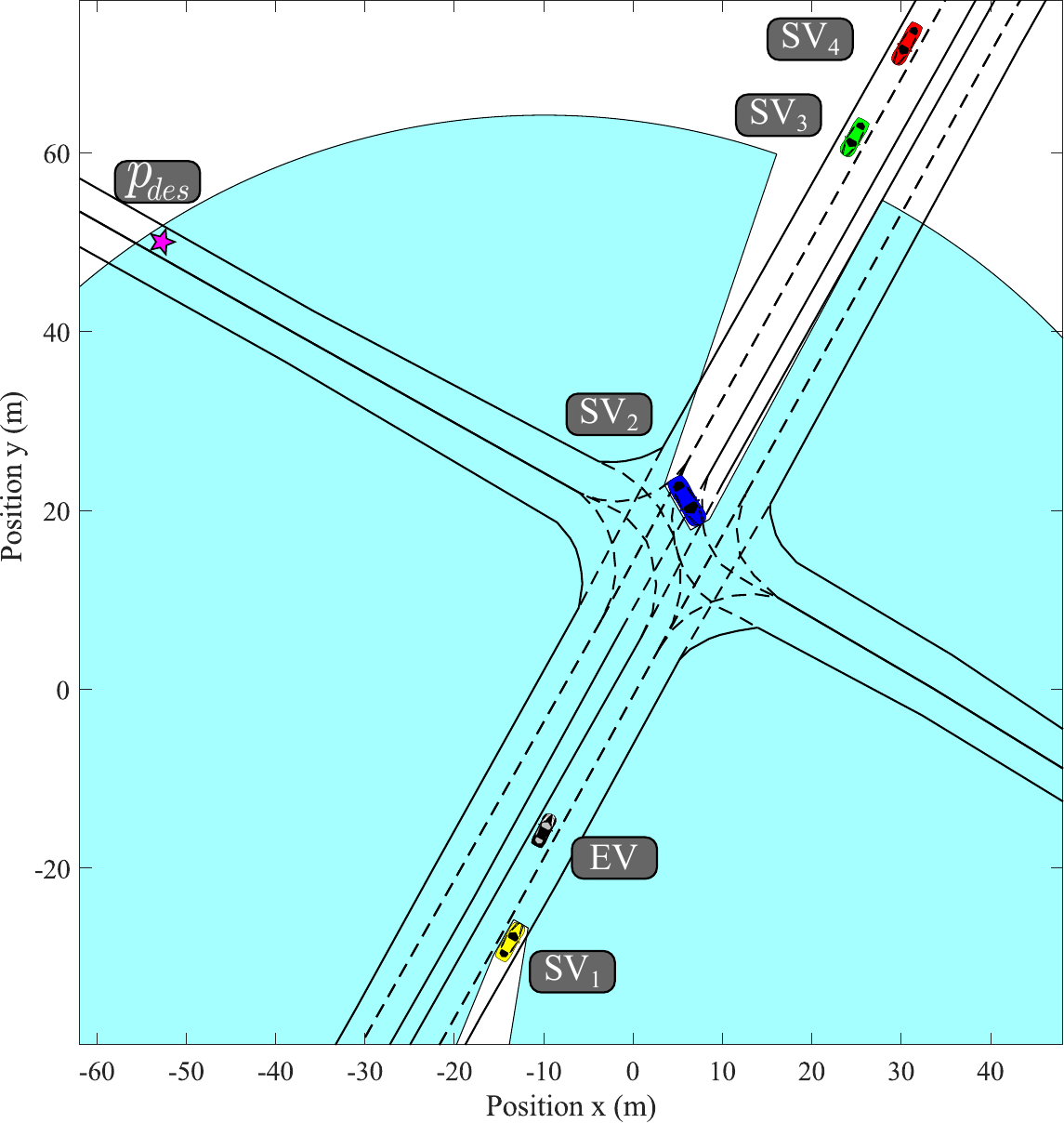}
    \caption{Intersection scenario layout. The cyan area represents the FOV of the EV. Vehicles outside the FOV are not perceived by the EV.}
    \label{fig: initialPos_occlusions_intersection}
\end{figure}

\begin{figure}[!ht]
\centering
\subfloat[]{\includegraphics[width=\linewidth]{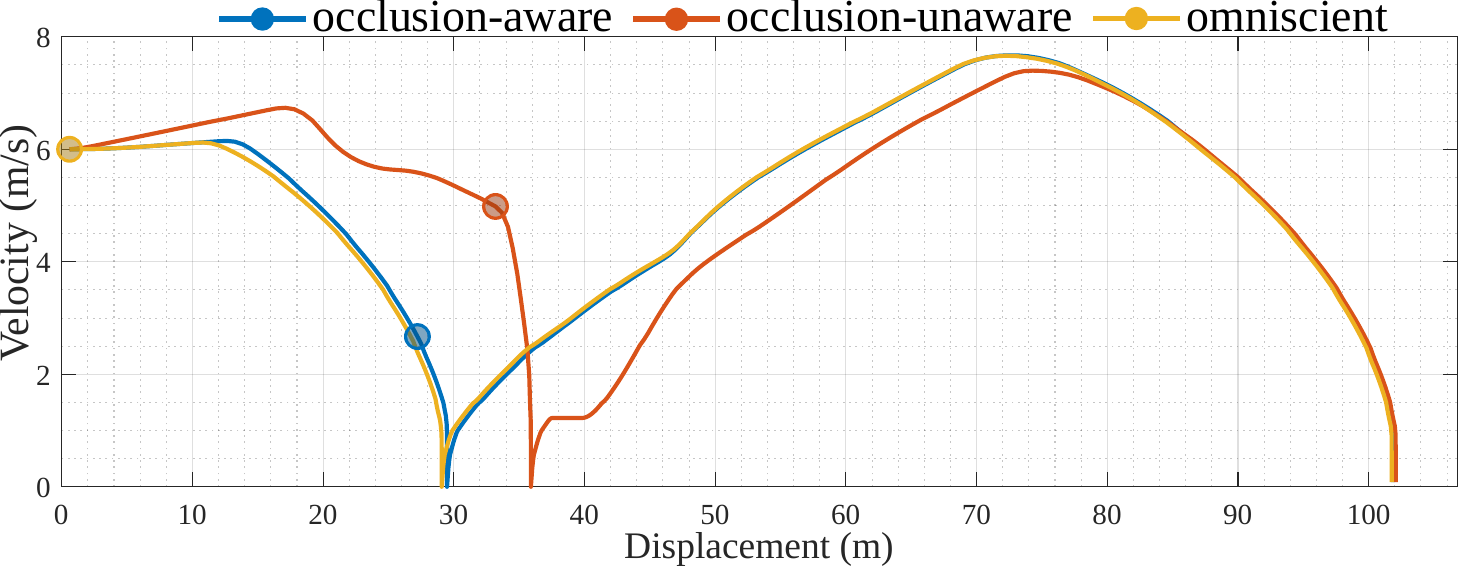}%
\label{fig: vel_intersection}}

\subfloat[]{\includegraphics[width=\linewidth]{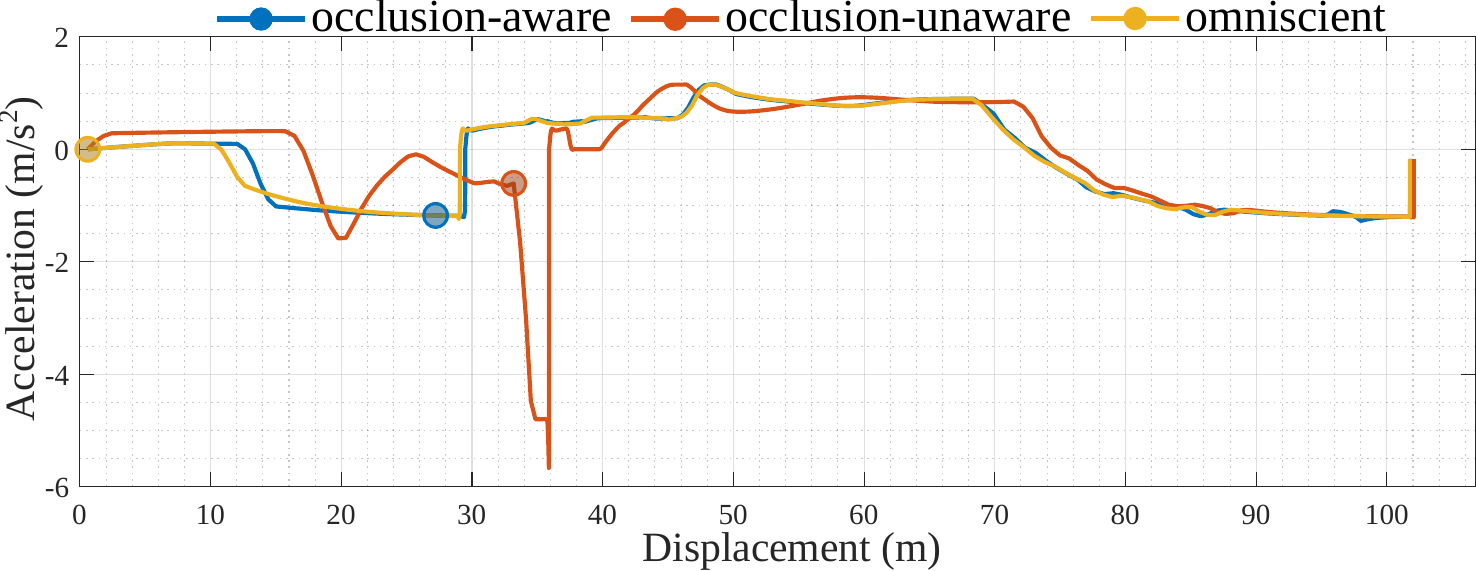}%
\label{fig: acc_intersection}}
\caption{Velocity and acceleration profiles for the different experiments on the intersection scenario. (a) Velocity profile. (b) Longitudinal acceleration profile.}
\label{fig: motion_intersection}
\vspace{-10pt}

\end{figure}

Another scenario where occlusions play a crucial role in the safety and comfort levels of the EV is presented in Fig.~\ref{fig: initialPos_occlusions_intersection}. In this situation, the EV is traveling towards an intersection where it wants to make a left turn. To do so, it must cross two lanes with right-of-way. Due to a stopped vehicle ($SV_2$) that also wants to make a left turn to the opposite side, the FOV of the EV is limited. The destination point $(p_{des})$ is $102~m$ away from the initial position of the EV.

Fig.~\ref{fig: motion_intersection} presents the velocity and acceleration over the path for the three experiments. Marked points are the moments that the EV perceives $SV_4$ for the first time. 

In the occlusion-aware experiment, although $SV_3$ and $SV_4$ are out of the FOV of the EV, the predictions generated by the HVs make the EV approach the intersection cautiously since the possible HV would have priority over it. A similar trajectory is planned for the omniscient experiment. On the other hand, in the occlusion-unaware experiment, since for the EV nobody is coming in the other direction, it plans a trajectory that slows down to cross but does not stop. At 5.8 s (marked point in Fig.~\ref{fig: motion_intersection}) within the initial trajectory, the EV perceives $SV_4$ and has to strongly break to avoid an imminent collision, making the EV stop 6.4 m after the other experiments. This can be seen in Fig.~\ref{fig: motion_intersection}, where the acceleration and velocity profiles over the path are displayed for the three experiments.

\begin{figure}[htpb]
    \centering
    \includegraphics[width=0.9\linewidth]{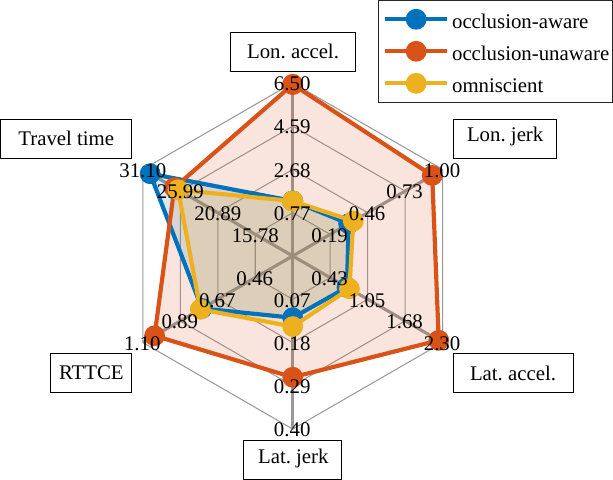}
    \caption{KPIs for the intersection experiment.}
    \label{fig: radar_occlusion_intersection}
\end{figure}

In this scenario, all experiments were able to reach the destination point. However, apart from the greater risk (KPI RTTCE), the EV faced higher values of acceleration and jerk in the occlusion-unaware experiment, resulting in lower comfort levels, as can be seen in Fig.~\ref{fig: radar_occlusion_intersection}. Regarding the travel time, the occlusion-aware experiment arrived at the destination point 3.8 s later than the omniscient experiment. The reason for that is that the EV with the motion grid provided by the former waits until the FOV obstructed by $SV_2$ is clear to start its movement, whereas the latter has full awareness of the scene and knows in advance that it can safely cross the intersection.

\vspace{-5pt}

\section{Concluding remarks}
\label{sec: conclusion}

In this work, the MultIAMP framework has been extended to tackle occlusions caused by obstructed view or sensor range. The implementation was evaluated in two simulated scenarios, where results showed that the occlusion-aware MultIAMP enables a similar performance of a motion planner where the prediction is considered omniscient.

Starting from an interaction-aware approach for prediction, the contributions of this paper introduce capabilities to manage occlusions, thereby enhancing prediction performance. In addition to demonstrating an increased ability to handle more complex scenarios involving occlusions, the results show similar planner performance when assuming omniscience compared to employing occlusion-aware techniques.

As future work, the framework will be extended and integrated with a state-of-the-art perception system~\cite{jimenez_TIV_2023} in order to evaluate its capabilities with real perception information.

\bibliographystyle{IEEEtran}
\bibliography{IEEEabrv,bibfile}

\begin{thebibliography}{10}
\providecommand{\url}[1]{#1}
\csname url@samestyle\endcsname
\providecommand{\newblock}{\relax}
\providecommand{\bibinfo}[2]{#2}
\providecommand{\BIBentrySTDinterwordspacing}{\spaceskip=0pt\relax}
\providecommand{\BIBentryALTinterwordstretchfactor}{4}
\providecommand{\BIBentryALTinterwordspacing}{\spaceskip=\fontdimen2\font plus
\BIBentryALTinterwordstretchfactor\fontdimen3\font minus \fontdimen4\font\relax}
\providecommand{\BIBforeignlanguage}[2]{{%
\expandafter\ifx\csname l@#1\endcsname\relax
\typeout{** WARNING: IEEEtran.bst: No hyphenation pattern has been}%
\typeout{** loaded for the language `#1'. Using the pattern for}%
\typeout{** the default language instead.}%
\else
\language=\csname l@#1\endcsname
\fi
#2}}
\providecommand{\BIBdecl}{\relax}
\BIBdecl

\bibitem{Orzechowski_2018}
P.~F. Orzechowski, A.~Meyer, and M.~Lauer, ``Tackling occlusions \& limited sensor range with set-based safety verification,'' in \emph{2018 21st International Conference on Intelligent Transportation Systems (ITSC)}, 2018, pp. 1729--1736.

\bibitem{Neel_2020}
G.~Neel and S.~Saripalli, ``Improving bounds on occluded vehicle states for use in safe motion planning,'' in \emph{2020 IEEE International Symposium on Safety, Security, and Rescue Robotics (SSRR)}, 2020, pp. 268--275.

\bibitem{Sánchez_2022}
J.~M.~G. Sánchez, T.~Nyberg, C.~Pek, J.~Tumova, and M.~Törngren, ``Foresee the unseen: Sequential reasoning about hidden obstacles for safe driving,'' in \emph{2022 IEEE Intelligent Vehicles Symposium (IV)}, 2022, pp. 255--264.

\bibitem{Yu_2019}
M.-Y. Yu, R.~Vasudevan, and M.~Johnson-Roberson, ``Occlusion-aware risk assessment for autonomous driving in urban environments,'' \emph{IEEE Robotics and Automation Letters}, vol.~4, no.~2, pp. 2235--2241, 2019.

\bibitem{Wang_2022}
D.~Wang, W.~Fu, Q.~Song, and J.~Zhou, ``Potential risk assessment for safe driving of autonomous vehicles under occluded vision,'' \emph{Scientific Reports}, vol.~12, 03 2022.

\bibitem{Hubmann_2019}
C.~Hubmann, N.~Quetschlich, J.~Schulz, J.~Bernhard, D.~Althoff, and C.~Stiller, ``A pomdp maneuver planner for occlusions in urban scenarios,'' in \emph{2019 IEEE Intelligent Vehicles Symposium (IV)}, 2019, pp. 2172--2179.

\bibitem{Zhang_2021}
C.~Zhang, F.~Steinhauser, G.~Hinz, and A.~Knoll, ``Improved occlusion scenario coverage with a pomdp-based behavior planner for autonomous urban driving,'' in \emph{2021 IEEE International Intelligent Transportation Systems Conference (ITSC)}, 2021, pp. 593--600.

\bibitem{Wang_2023}
D.~Wang, W.~Fu, J.~Zhou, and Q.~Song, ``Occlusion-aware motion planning for autonomous driving,'' \emph{IEEE Access}, vol.~11, pp. 42\,809--42\,823, 2023.

\bibitem{Bouton_2019}
M.~Bouton, A.~Nakhaei, K.~Fujimura, and M.~J. Kochenderfer, ``Safe reinforcement learning with scene decomposition for navigating complex urban environments,'' in \emph{2019 IEEE Intelligent Vehicles Symposium (IV)}, 2019, pp. 1469--1476.

\bibitem{Kamran_2020}
D.~Kamran, C.~F. Lopez, M.~Lauer, and C.~Stiller, ``Risk-aware high-level decisions for automated driving at occluded intersections with reinforcement learning,'' in \emph{2020 IEEE Intelligent Vehicles Symposium (IV)}, 2020, pp. 1205--1212.

\bibitem{vinicius_ACCESS}
V.~Trentin, A.~Artuñedo, J.~Godoy, and J.~Villagra, ``Interaction-aware intention estimation at roundabouts,'' \emph{IEEE Access}, vol.~9, pp. 123\,088--123\,102, 2021.

\bibitem{vinicius_TIV}
------, ``Multi-modal interaction-aware motion prediction at unsignalized intersections,'' \emph{IEEE Transactions on Intelligent Vehicles}, vol.~8, no.~5, pp. 3349--3365, 2023.

\bibitem{thesis_Victor}
\BIBentryALTinterwordspacing
V.~Jimenez~Bermejo, ``Grid-based perception framework using lidar sensors : a multi-representation approach,'' January 2024, unpublished. [Online]. Available: \url{https://oa.upm.es/80767/}
\BIBentrySTDinterwordspacing

\bibitem{Wang_2021}
L.~Wang, C.~Burger, and C.~Stiller, ``Reasoning about potential hidden traffic participants by tracking occluded areas,'' in \emph{2021 IEEE International Intelligent Transportation Systems Conference (ITSC)}, 2021, pp. 157--163.

\bibitem{Juan_IV2}
J.~F. Medina-Lee, J.~Godoy, A.~Artuñedo, and J.~Villagra, ``Speed profile generation strategy for efficient merging of automated vehicles on roundabouts with realistic traffic,'' \emph{IEEE Transactions on Intelligent Vehicles}, pp. 1--15, 2022.

\bibitem{juan_sensors_2021}
J.~F. Medina-Lee, A.~Artu{ñ}edo, J.~Godoy, and J.~Villagra, ``Merit-based motion planning for autonomous vehicles in urban scenarios,'' \emph{Sensors}, vol.~21, no.~11, pp. 1--25, 2021.

\bibitem{eggert_2017}
J.~Eggert and T.~Puphal, ``Continuous risk measures for adas and ad,'' in \emph{Proceedings Future Active Safety Technology Symposium}, 2017, pp. 1--8.

\bibitem{vinicius_TIV2}
J.~F. Medina-Lee, V.~Trentin, J.~L. Hortelano, A.~Artuñedo, J.~Godoy, and J.~Villagra, ``$ia(mp)^{2}$: Framework for online motion planning using interaction-aware motion predictions in complex driving situations,'' \emph{IEEE Transactions on Intelligent Vehicles}, pp. 1--16, 2023.

\bibitem{Juan_IV1}
J.~F. Medina-Lee, A.~Artuñedo, J.~Godoy, and J.~Villagra, ``Reachability estimation in dynamic driving scenes for autonomous vehicles,'' in \emph{2020 IEEE Intelligent Vehicles Symposium (IV)}, 2020, pp. 2133--2139.

\bibitem{jimenez_TIV_2023}
V.~Jiménez, A.~A. J.~Godoy~and, and J.~Villagra, ``Object-level semantic and velocity feedback for dynamic occupancy grids,'' \emph{IEEE Transactions on Intelligent Vehicles}, vol.~8, no.~7, pp. 3936--3953, 2023.

\end{thebibliography}
\end{document}